\documentclass{isrpaper}

\usepackage{fontspec}

\usepackage{unicode-math}

\usepackage{orcidlink}

\usepackage{etoolbox}
\AtBeginEnvironment{tabular}{\vspace{3pt}\scriptsize}
\usepackage[round,authoryear]{natbib}

\newcommand{\parencite}[1]{\citep{#1}}
\newcommand{\textcite}[1]{\citet{#1}}

\title{Cross-Model Distillation of a Human-Pose Foundation Model from Unannotated Infant Video for Markerless 3D Pose Estimation}

\author{
R. James Cotton\,\orcidlink{0000-0001-5714-1400}\,$^{1,2}$, Divya Joshi$^{1,3}$, Colleen Peyton$^{3,4}$}

\affiliations{
$^{1}$Shirley Ryan AbilityLab \\ $^{2}$Department of Physical Medicine and Rehabilitation, Northwestern University \\ $^{3}$Department of Physical Therapy and Human Movement Science, Northwestern University \\ $^{4}$Department of Pediatrics, Northwestern University}

\paperdate{September 2026}

\correspondence{rcotton@sralab.org}

\paperabstract{Spontaneous movement is one of the earliest windows onto an infant's neuromotor health, and structured clinical instruments that score it are validated early predictors of cerebral-palsy risk. However, they require specially trained raters, are time-consuming, and carry inter-rater variability. This motivates automated, video-based markerless assessment, especially as marker-based motion capture is impractical in infants. Yet the foundation models that make markerless capture possible are trained almost entirely on adults: our recent multi-view infant study found that no single model is jointly best, with strong 2D keypoint accuracy and direct 3D body recovery split across different models \parencite{joshi_markerless_2026}. While that study identifies this trade-off, it does not resolve it. Here, we perform cross-model distillation from the Sapiens 2 pose model into the SAM 3D Body model, using unannotated infant video alone. A frozen teacher supplies dense pseudo-labels, and a differentiable renderer aligns the predicted mesh to them in the training loop. On eleven held-out infants (18 sessions, 173 recordings) under our prior study's multi-view protocol, fine-tuning improves same-view 2D keypoint agreement with the Sapiens reference (median body percentage of correct keypoints @ 10px 0.22\rightarrow0.42, face 0.22\rightarrow0.42) and Procrustes-aligned mean per joint 3D position error (25.5\rightarrow22.2 mm). This demonstrates how cross-model distillation improves SAM 3D Body model performance on infants.}

\begin{document}

\maketitle

\section{Introduction}

Movement in early infancy is a direct window onto the developing nervous system, and the quality of an infant's spontaneous movement is among the earliest observable signs of motor impairment \parencite{Novak2017, Herskind2015EarlyPalsy}. Clinicians exploit this through structured observation using validated instruments, such as the Prechtl General Movements Assessment \parencite{Bosanquet2013AChildren, Einspeiler2005}, the Test of Infant Motor Performance \parencite{Spittle2008ALife}, and the Baby Observational Selective Control AppRaisal \parencite{Peyton2024BabyYears, SukalMoulton2024BabyPalsy, Barbosa2024BabyPerformance, Peyton2024BabyLongitudinal}, to predict later motor outcomes including cerebral palsy. However, these observational assessments depend on specially trained raters, are time-consuming, and carry inter-rater variability that limits their reach in routine care \parencite{joshi_markerless_2026, Noble2012NeonatalReview, Bosanquet2013AChildren}. This motivates automated, video-based assessment that quantifies infant movement at scale. Markerless motion capture meets this need: it is non-invasive, scalable, and feasible in infants, a population in which affixing markers is impractical and often contraindicated \parencite{joshi_markerless_2026}. This is already clinically actionable: kinematic features computed from 2D keypoints in single-camera infant video predict clinician-assessed General Movements Assessment scores in a large preregistered cohort of at-risk infants, using keypoint detectors never optimized for infants \parencite{segado_assessing_2024}.

The clinically meaningful signal is not sparse 2D keypoints, but rather dense, whole-body 3D kinematics: the joint angles that describe how an infant actually moves, and whether individual joints move independently of one another --- the selective motor control \parencite{Cahill-Rowley2014EtiologyPalsy} whose emergence in infancy predicts later cerebral palsy \parencite{Peyton2024BabyLongitudinal}. Recovering these motion quantities from raw video pixels requires a 3D body model \parencite{cotton_differentiable_2025}. Building on the SMPL-style class of surface mesh models for humans \parencite{loper_smpl_2015, pavlakos_smplx_2019}, we use the Momentum Human Rig (MHR) \parencite{ferguson_mhr_2025}, which decouples a posable skeleton from a shape-dependent surface. Foundation models now regress these parameters directly from a single image: SAM 3D Body \parencite{yang_sam_2025} predicts MHR parameters and in our prior study showed promising performance for infants \parencite{joshi_markerless_2026}. However, models such as Sapiens and Sapiens 2 \parencite{khirodkar_sapiens_2024, khirodkar_sapiens2_2026} set the state of the art for 2D keypoint accuracy while not estimating joint angles.

These models are trained primarily on adults, who appear different visually, pose differently, and have different body proportions \parencite{Sciortino2017InfantProportions, huang2021invariant, bose2025SHIFT}. Furthermore, labeled infant pose data is scarce \parencite{GamaAutomaticMethods}, and marker-based ground truth is infeasible \parencite{joshi_markerless_2026}. Infant-specific shape spaces exist, such as the Skinned Multi-Infant Linear Model (SMIL) \parencite{hesse_smil_2018} and the lifespan model Anny \parencite{bregier_anny_2025}, but lack foundation models for predicting those parameters from images.

Our recent work benchmarked MeTRAbs-ACAE \parencite{sarandi_learning_2023}, SAM 3D Body \parencite{yang_sam_2025}, and Sapiens \parencite{khirodkar_sapiens_2024, khirodkar_sapiens2_2026} on multi-view infant recordings \parencite{joshi_markerless_2026}. Sapiens achieved the best 2D multi-view consistency but produced no direct 3D body estimates; SAM 3D Body produced fairly good 3D kinematics, but its predicted keypoints were not as geometrically consistent as those of Sapiens.

Here, we seek to close this gap by distilling pseudo-labels from a strong Sapiens 2 teacher into the SAM 3D Body model to improve predictions on infants, using only unlabeled infant video. A frozen Sapiens 2 teacher supplies dense pseudo-labels of keypoints, depth, surface normals, and silhouettes, and a differentiable renderer aligns the student's predicted mesh to them in the loop. On held-out infants this improves both 2D agreement and Procrustes-aligned 3D error.

\section{Methods}

\begin{figure}[!htbp]
\centering
\includegraphics[width=1\linewidth]{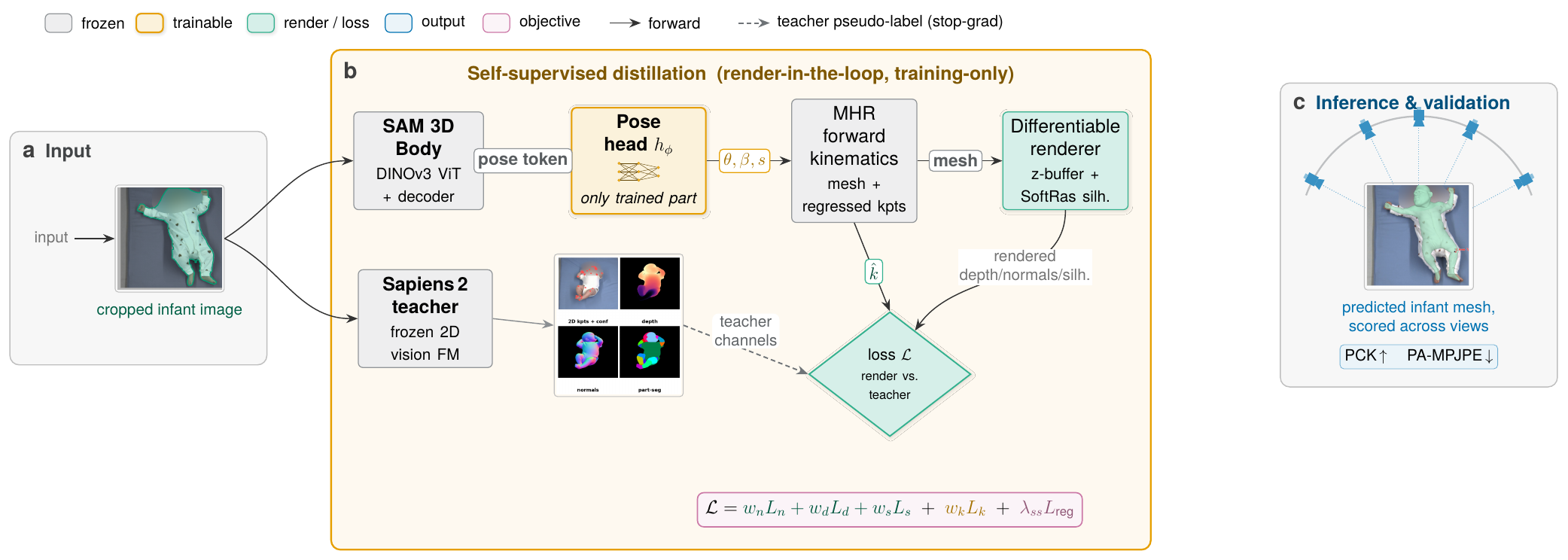}
\caption[]{Method overview. \textbf{(a)} A per-frame infant crop is the shared input to both teacher and student. \textbf{(b)} Render-in-the-loop
cross-model distillation (training only): the frozen SAM 3D Body backbone and decoder feed a small pose head
$h_\phi$ --- the only trained part --- whose MHR parameters drive differentiable forward kinematics and a
differentiable renderer. A frozen Sapiens 2 teacher labels the same crop with dense pseudo-labels
(2D keypoints, depth, surface normals, part segmentation); the loss $\mathcal{L}$ compares the render
against those channels, backpropagating into the pose head alone. \textbf{(c)} At inference the adapted
model runs per camera, validated against our prior study's multi-view benchmark.}
\label{fig-overview}
\end{figure}

\textbf{Overview.} We adapt a strong-3D human-pose foundation model to infants by distilling a strong-2D foundation-model teacher into it, using only unannotated infant video. The frozen teacher produces dense per-frame pseudo-labels, and a small trainable pose head is updated so that a differentiable render of the predicted body matches the teacher's pseudo-labels. We then re-run our established multi-view infant validation protocol \parencite{joshi_markerless_2026} to quantify the gain (Figure~\ref{fig-overview}).

\subsection{Teacher, student, and the trainable surface}

The teacher is Sapiens 2, a vision-transformer human-vision foundation model: it emits 2D keypoints with confidence scores, metric depth, surface normals, and body-part segmentation \parencite{khirodkar_sapiens2_2026}. The infant crop and mask come from SAM 3 \parencite{carion2026sam}, prompted with the text ``baby'' so that the subject is chosen by concept rather than by size or position --- the caregiver is often the larger and more central person in these recordings --- and we keep only its highest-scoring detection. A frame is accepted only if that detection scores above 0.5, covers at least 5000 pixels, and does not run off the frame edge; after the teacher has run, the frame is also dropped unless the infant mask agrees with the teacher's own body segmentation, at an intersection-over-union above 0.35 or with at least 70\% of the infant mask falling inside it. The teacher runs live during training, labelling frames on the fly. Its 2D keypoints follow the Goliath-308 layout: the first 70 are body and hand joints, and the remainder are dense facial landmarks \parencite{khirodkar_sapiens_2024, khirodkar_sapiens2_2026}. Distillation is supervised by these keypoints and their confidence scores, together with the depth and normal maps. The body-part segmentation and the separate infant mask serve a gating role, controlling which pixels and keypoints each loss term is allowed to see.

The student is SAM 3D Body \parencite{yang_sam_2025}, which recovers a Momentum Human Rig (MHR) body \parencite{ferguson_mhr_2025} from a single top-down crop. A DINOv3 vision-transformer backbone encodes the crop, a promptable transformer decoder condenses it into a pose token, and a small feed-forward pose head projects that token to the MHR parameter vector: global orientation, body pose, per-hand PCA coefficients, facial expression, and identity shape (45-dim) and skeletal scale (28-dim) coefficients. Differentiable MHR forward kinematics --- which decouples kinematic-tree scaling from pose --- then maps these parameters to a posed body mesh and its regressed keypoints,  which are the two outputs the distillation objective supervises. The mesh is passed to the renderer to produce depth, normal, and silhouette maps; the regressed keypoints follow the Goliath-308 layout (the first 70 are MHR body-and-hand joints, the remainder dense facial landmarks), matching the teacher so that all 308 keypoints can be supervised directly.

Distillation updates only the pose head's projection network --- a two-layer feed-forward block of a few hundred thousand parameters --- while the backbone, decoder, and MHR forward kinematics stay fixed. Within the pose head, the gradient is masked on the rows of the final linear layer that emit the 45 shape and 28 scale coefficients, so those output weights stay at their pretrained values while the rest of the head trains. The predicted shape and scale are not thereby held constant, since the features feeding those rows still adapt; what is removed is the direct route from the pixel losses into the shape readout. In the accompanying Supplementary Material, those rows are instead trained under an L2 prior $\lambda_{ss}$, swept in strength, and the comparison motivating the masked configuration is reported. The entire render-in-the-loop pipeline --- teacher, student, renderer, and forward kinematics --- is differentiable end-to-end on a MuJoCo-compatible JAX/Equinox stack \parencite{kidger_equinox_2021, todorov_mujoco_2012, jax_2018}.

\subsection{Render-in-the-loop objective}

For each accepted frame the student's pose head predicts the posed body from a fixed top-down crop. The predicted mesh is rendered differentiably and compared against the teacher's channels. Depth and surface normals are rendered by barycentric interpolation of the mesh under a z-buffered rasterizer, so gradients from those losses flow back through the render into the pose head.  The silhouette is rendered with a soft rasterizer: per-pixel coverage is the largest of Gaussian kernels centred on the projected visible vertices, so it varies smoothly with vertex position and the silhouette loss is differentiable with respect to the outline. Back-face and depth culling remain a hard, detached test, so only the coverage field carries gradient. Per-vertex normals are mapped into the teacher's convention before comparison.  These rendered outputs are compared against the teacher pseudo-labels through a combined objective,

\begin{equation}
\mathcal{L} =
w_{\text{kpt}}\,L_{\text{kpt}}
+ w_{\text{normal}}\,L_{\text{normal}}
+ w_{\text{depth}}\,L_{\text{depth}}
+ w_{\text{silh}}\,L_{\text{silh}}
+ \lambda_{ss}\,L_{\text{reg}},
\end{equation}

in which the keypoint, normal, and depth terms are active and the silhouette term is weighted lightly.

The keypoint term $L_{\text{kpt}}$ is a confidence-weighted robust loss on the 2D keypoints that separates major body joints, head anchors, and dense face landmarks, weighting the major joints and head anchors fully but the dense face landmarks an order of magnitude less. The teacher supplies 238 dense face landmarks against 12 major body joints and 5 coarse head anchors (nose, eyes and ears), so at equal weight the face --- which is not the movement we set out to recover --- would dominate the residual; down-weighting it produces clean, baby-sized fits. Teacher keypoints are also gated: those below the confidence threshold are dropped, and each remaining keypoint is looked up in the infant mask at its own pixel and given zero weight if it lands outside. This gating matters most for the caregiver, who is frequently in frame or in contact with the infant, so a hand or arm the teacher detects on the caregiver is excluded rather than fit, along with keypoints that fire on blankets and background clutter. The normal term $L_{\text{normal}}$ is a cosine distance between rendered and teacher surface normals over an eroded body mask. Because the teacher predicts only relative monocular depth, the depth term $L_{\text{depth}}$ is a per-frame affine-aligned L1: a closed-form scale-and-offset fit reconciles the two depth conventions and absorbs the teacher-versus-MHR camera-frame offset before comparison. The silhouette term $L_{\text{silh}}$ aligns the rendered mesh boundary to the infant mask bidirectionally, retracting it where the mask excludes the body and extending it where the mask includes it. The Sapiens face/neck and hair segmentation classes are removed from the normal and depth losses and the silhouette term is disabled inside that region, since the adult-mean blendshapes cannot reach infant head proportions and pressure from the dense losses there deforms the body rather than the head. The regularizer $L_{\text{reg}}$ is an L2 prior on shape and scale, inactive in the reported model because its shape and scale rows are masked; the sweep in the accompanying Supplementary Material document unfreezes them and varies $\lambda_{ss}$. All terms are differentiable back through the MHR forward kinematics.

\subsection{Online data pipeline}

The loader interleaves many recordings at once, so consecutive samples come from different infants, and within a recording it samples at a fixed stride of ten frames in short temporally-adjacent runs rather than independently. A cap on how many frames a recording may contribute before it is set aside forces a run to cover the cohort rather than the handful of recordings it happens to open with. A held-out tenth of the corpus, split by subject so that all timepoints of an infant fall on the same side, supplies an in-loop validation loss; every reported accuracy number comes from the separate multi-camera cohort described below.

\subsection{Infant video corpus}

The training corpus is a large set of raw, unannotated single-camera infant recordings. The reported cohort comprises 543 recordings of 311 unique infants whose shorter side is at least 720 px, skewing preterm (median gestational age 31 weeks, IQR 27--38; median post-menstrual age at recording 52 weeks) --- Table~\ref{tbl-cohort-demographics} summarizes the cohort demographics. The eleven validation infants are not part of this monocular corpus; they are additionally excluded at index build, so the cohort is leakage-free.

The eleven validation infants were recorded on the multi-camera system of our prior study \parencite{joshi_markerless_2026}: eight synchronized RGB cameras (FLIR BlackFly S, f/1.4 lenses) at 29 frames per second, on tripods surrounding a mat with the infant supine at its centre, each session comprising 8--10 trials of roughly a minute of spontaneous movement. Intrinsics and extrinsics are calibrated per session from a moving checkerboard or ChArUco target, and the 3D reference is a robust multi-view triangulation that down-weights geometrically inconsistent detections \parencite{Roy2022OnEstimation} and low-confidence ones \parencite{cotton_improved_2023}.

The parents and/or guardians of all participants provided informed, written consent to participate in this study, which was approved by the institutional review board of Northwestern University.

\begin{table}
\centering
\caption[]{Training-corpus demographics --- the unlabeled monocular distillation cohort (543 recordings of 311 infants). Post-menstrual age is per recording; the other rows are per infant. The diagnosis row groups the clinical labels into \textit{likely CP} (confirmed CP, 129, plus likely CP, 15) versus \textit{unlikely CP} (typically developing, 157, plus unlikely CP, 8). Parenthetical counts give the denominator where a field was not recorded for every infant.}

\begin{tabular}{p{\dimexpr 0.500\linewidth-2\tabcolsep}p{\dimexpr 0.500\linewidth-2\tabcolsep}}
\toprule
Characteristic & Value \\
\hline
Infants (unique) & 311 \\
Recordings & 543 \\
Recordings per infant, median [range] & 1 [1--13] \\
Post-menstrual age at recording (wk), median [IQR], range & 52 [42--54], 28--61 \\
Gestational age (wk), median [IQR], range & 31 [27--38], 22--53 (n = 305) \\
Preterm : term & 204 : 105 (2 unrecorded) \\
Sex (M : F) & 174 : 132 (5 unrecorded) \\
Diagnosis (likely CP : unlikely CP) & 144 : 165 (2 unrecorded) \\
\bottomrule
\end{tabular}
\label{tbl-cohort-demographics}
\end{table}

\subsection{Implementation details}

Only the pose head weights are optimized ---  excluding the masked shape and scale readout rows --- using Adam with global-norm gradient clipping (clip 1.0, no weight decay). On a 40 GB GPU this system could only fit a batch size of two. The reported objective weights are $(w_{\text{kpt}}, w_{\text{normal}}, w_{\text{depth}}, w_{\text{silh}}) = (0.5, 0.5, 0.5, 0.1)$ with $\lambda_{ss}=0$; keypoint sub-weights are 1.0 for major joints and head anchors and 0.1 for dense face landmarks, gated at teacher confidence 0.3. We train on the cohort above with the per-video coverage quota, photometric and crop-scale augmentation, and a peak learning rate of $2\times10^{ -5}$ reached by a short linear warmup and then held constant, for roughly 20k steps.

\subsection{Evaluation protocol}

We evaluate on that held-out multi-camera cohort using two readouts.

The first is per-camera 2D agreement: the student's 2D keypoints are scored against the Sapiens 2D keypoints in the \textit{same} camera view, isolating the distillation gain from a monocular perspective. We report the percentage of correct keypoints (PCK) at fixed pixel radii (2, 5, and 10 px) over the body, face, and all-keypoint groups, under the same teacher-confidence gate used in training. Same-view metrics are computed on the earliest recording of each infant so that every infant contributes equally.

The second is PA-MPJPE: the student's monocular 3D keypoints are compared against a multi-view triangulated \parencite{Roy2022OnEstimation} reference under a per-frame Procrustes alignment with scale, making the metric invariant to the global similarity transform a monocular setup cannot recover. The alignment is solved over eleven reliably triangulated body joints (shoulders, elbows, wrists, knees, ankles, and neck as root); we additionally report a thirteen-joint variant that includes the hips, which are otherwise excluded due to poor triangulation in occluded infants. Both the 2D same-view targets and the 3D triangulated reference are derived from Sapiens 2D estimates across the calibrated multi-camera rig, following our prior study's protocol \parencite{joshi_markerless_2026} --- they are model-derived references rather than marker-based ground truth, so both readouts measure agreement with a strong automatic reference, not an independent gold standard.

All frames are sampled at a fixed stride of ten before scoring. We report the median across infants with a normalized inter-quartile range ($\sigma_{\text{NIQR}} = 0.7413 \cdot \text{IQR}$) as an outlier-robust spread, with mean and standard deviation reported alongside, and assess significance with a paired Wilcoxon signed-rank test across the eleven infants.

Both readouts are reported in Table~\ref{tbl-main} alongside our prior study's multi-view benchmark harness --- reprojection error, geometric consistency, and a broader-keypoint PA-MPJPE --- rerun on the same held-out infants for direct comparison \parencite{joshi_markerless_2026}.

Geometric consistency $\text{GC}_d$ \parencite{cotton_improved_2023} is the fraction of detected 2D keypoints that a reconstruction reprojects to within $d$ pixels, counting only detections whose confidence exceeds $\lambda$. Writing $\delta_{t,j,c} = \lVert \Pi_c \mathbf{x}_{t,j} - y_{t,j,c} \rVert$ for the image-plane distance in camera $c$ between the triangulated 3D keypoint $\mathbf{x}_{t,j}$ reprojected through that camera's projection operator $\Pi_c$ (including intrinsic distortion) and the corresponding 2D detection $y_{t,j,c}$ of confidence $w_{t,j,c}$,

\begin{equation}
\text{GC}_d = \frac{\sum_{t,j,c} \mathbb{1}\!\left(\delta_{t,j,c} < d\right) \mathbb{1}\!\left(w_{t,j,c} > \lambda\right)}{\sum_{t,j,c} \mathbb{1}\!\left(w_{t,j,c} > \lambda\right)},
\end{equation}

with $\lambda = 0.5$ and, following our prior study, non-facial keypoints only. We report $\text{GC}_5$ and $\text{GC}_{10}$; higher is better.

\section{Results}

We evaluate on 11 held-out multi-camera infants (18 capture sessions, 173 recordings), entirely separate from the distillation corpus by construction, sampling every recording at a fixed frame stride of ten. Table~\ref{tbl-main} consolidates all quantitative results into a single table, scored on the same strided frames of the same held-out infants, and groups metrics by their reference rather than by view protocol. The \textbf{multi-view consistency} rows replicate our previous paper's harness --- reprojection error and geometric consistency (GC) --- and carry the Sapiens teacher's published values for reference (\parencite{joshi_markerless_2026}). The \textbf{teacher-agreement} rows are our own readout: per-camera 2D PCK@10px against the Sapiens teacher's keypoints in the same view, and Procrustes-aligned 3D error (PA-MPJPE) over reliably triangulated core body joints, reported as the median (nIQR) across infants. Procrustes-aligned 3D error is reported over three nested keypoint sets. The first two are scored against a Sapiens-triangulated reference and the third against a reference triangulated from the MHR keypoints themselves, so the third is not directly comparable with the first two. The first is eleven core body joints --- shoulders, elbows, wrists, knees, ankles and neck --- and is our primary 3D metric. The second adds both hips, which the reference localizes poorly. The third is MHR-70, the model's complete keypoint output; 40 of those 70 are finger keypoints. The first two solve the alignment on the joints they score, the third over the full 70-keypoint layout.

\subsection{Qualitative results}

\begin{figure}[!htbp]
\centering
\includegraphics[width=0.8\linewidth]{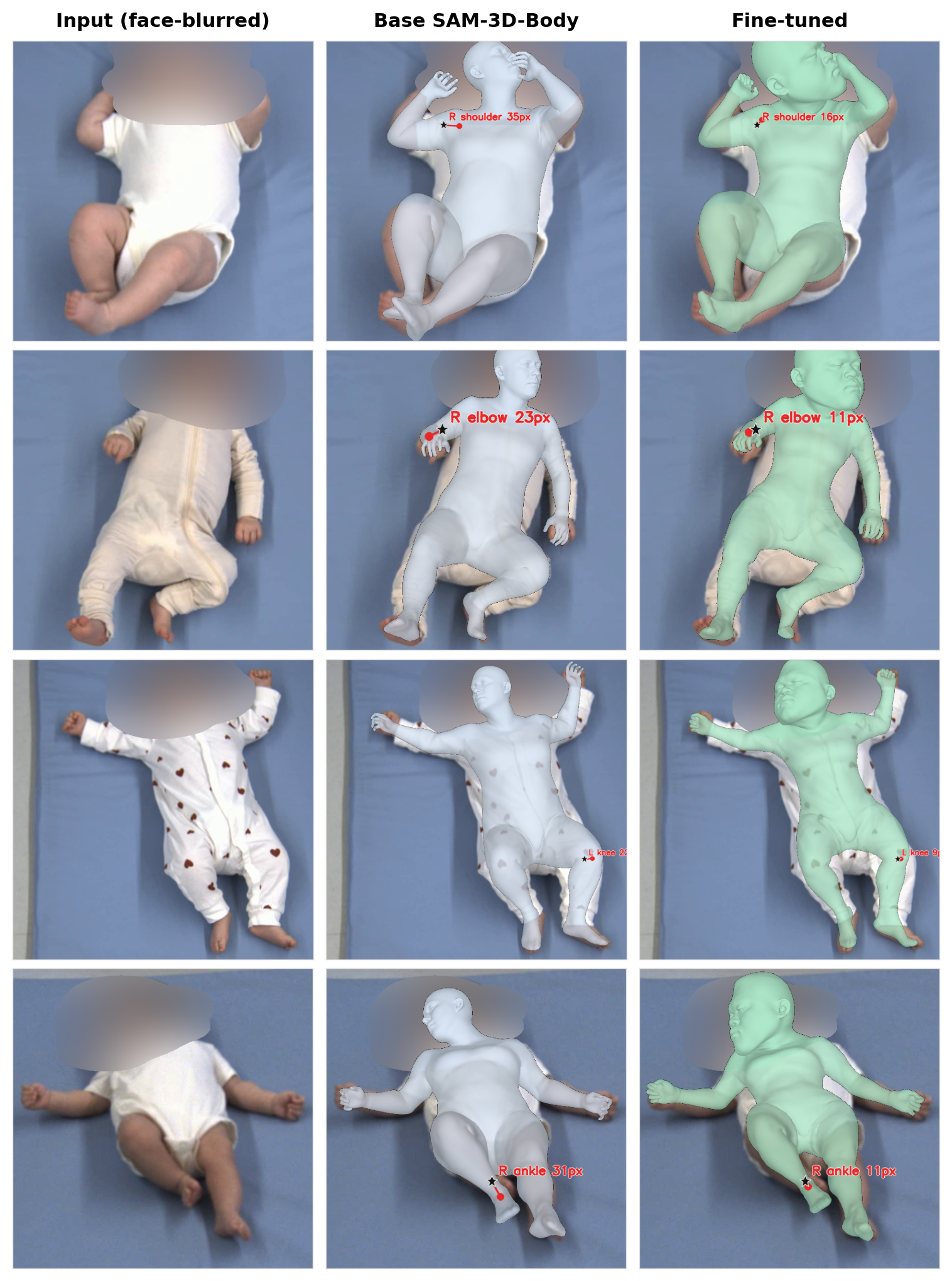}
\caption[]{Qualitative comparison on four held-out infants (rows), through the capture rig's
shared elevated camera. Columns: face-blurred input; base SAM-3D-Body mesh;
fine-tuned mesh. For each row we highlight the major limb joint where
fine-tuning most reduced the 2D distance to the Sapiens teacher keypoint for that
frame and camera; a black star marks the Sapiens target, and the line and label on
each mesh column give that model's pixel distance to it (fine-tuned consistently
closer). Targets are the Sapiens-Goliath 2D keypoints, computed as in our prior study.}
\label{fig-qualitative}
\end{figure}

Figure~\ref{fig-qualitative} compares the recovered MHR mesh between the base
SAM-3D-Body model and the fine-tuned student on four example frames from held-out infants, viewed from the rig's most elevated camera, which looks down on the mat. Fine-tuning consistently reduces the 2D distance to the Sapiens teacher keypoints, though the fitted mesh also exhibits shape artifacts --- an adult-proportioned face and an overly narrow, developed torso --- that persist after distillation.

\begin{figure}[!htbp]
\centering
\includegraphics[width=0.8\linewidth]{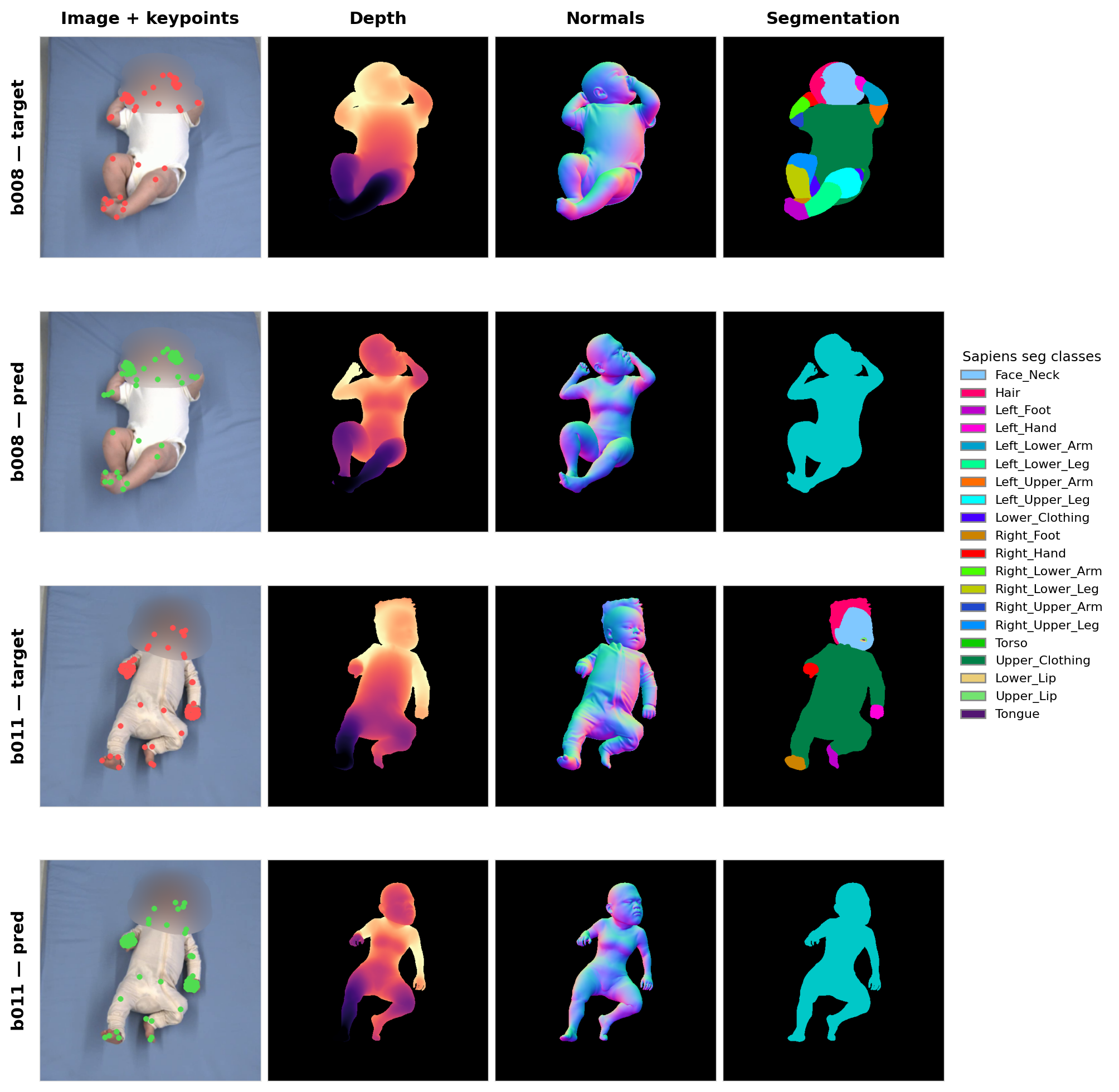}
\caption[]{Sapiens teacher pseudo-labels (target rows) vs fine-tuned MHR predictions (pred
rows) for two infants. Columns: 2D keypoints on the face-blurred photo, depth,
normals, segmentation. The segmentation channel was not used directly in supervision; only the binary infant mask was.}
\label{fig-channels}
\end{figure}

Figure~\ref{fig-channels} compares the Sapiens pseudo-label channels against the student's rendered predictions. The shape artifacts visible in Figure~\ref{fig-qualitative} are absent in the pseudo-labels themselves.

\subsection{Quantitative results}

\begin{figure}[!htbp]
\centering
\includegraphics[width=1\linewidth]{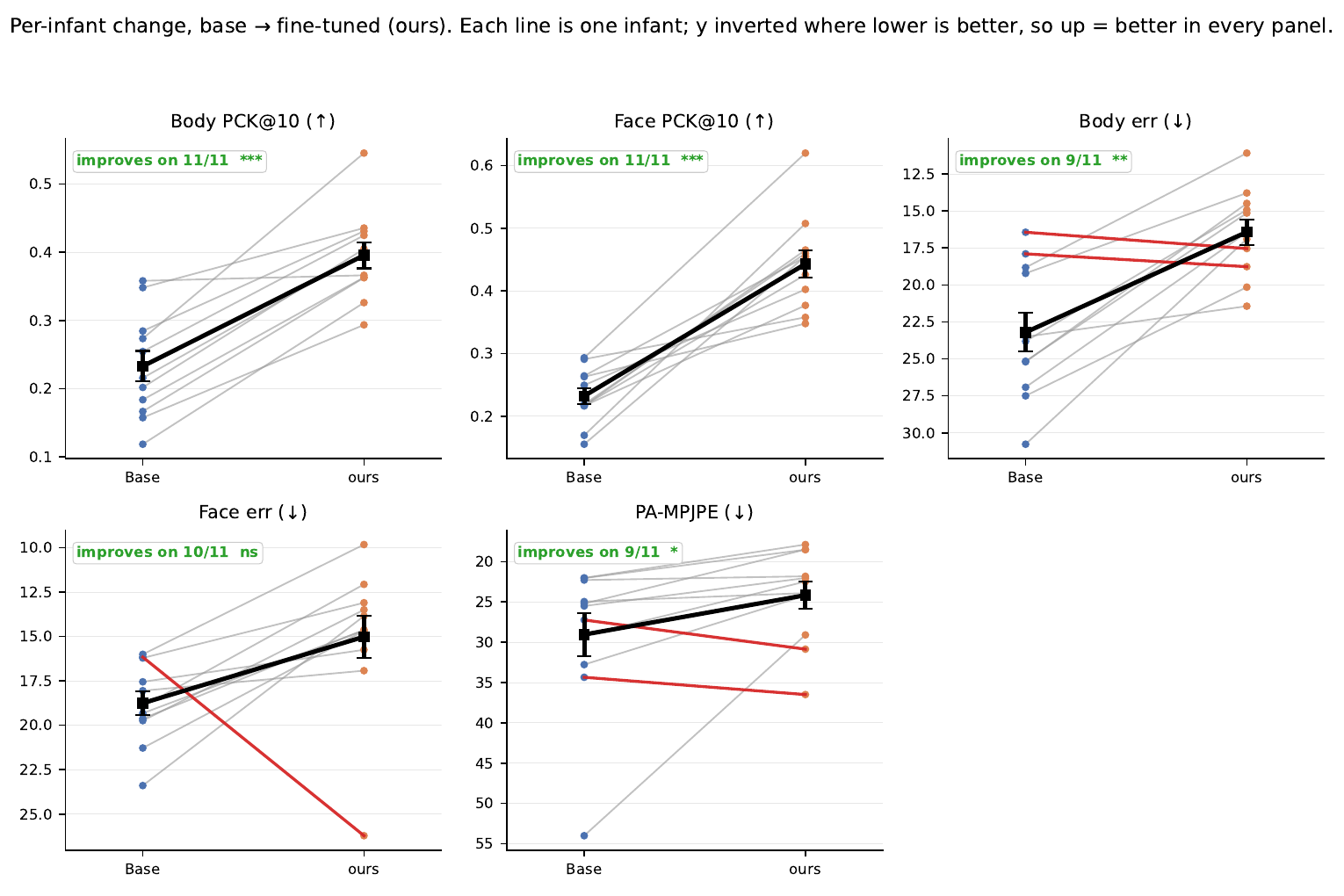}
\caption[]{Per-infant change, base \rightarrow fine-tuned, one line per infant with the heavy black
line the cohort mean ($\pm$ 1 SEM). Panels for lower-is-better metrics have the y-axis
inverted so ``up = better'' reads identically everywhere; regressing infants, where
any, are drawn in red --- the two PCK panels improve on all 11 infants, and the
pixel-error and PA-MPJPE panels have one to two regressions each.}
\label{fig-sameview-perinfant-slope}
\end{figure}

Distillation moves the student toward the Sapiens teacher (Table~\ref{tbl-main}); teacher-agreement metrics are reported as the median across infants with the normalized inter-quartile range ($\sigma_{\text{NIQR}} = 0.7413\cdot\text{IQR}$) as an outlier-robust spread. Keypoint agreement (PCK@10px, model vs. teacher in the same camera) approximately doubles --- body 0.216 \rightarrow 0.418 and face 0.219 \rightarrow 0.422 --- and the scale-invariant 3D PA-MPJPE over the reliably triangulated core body joints improves from 25.5 \rightarrow 22.2 mm, all significant under a paired Wilcoxon signed-rank test (p $\leq$ 0.005). Pixel error falls correspondingly (Figure~\ref{fig-sameview-perinfant-slope}).

We report PA-MPJPE primarily over the eleven body joints that exclude the hips, because the triangulated reference localizes infant hips unreliably due to frequent occlusions. Including the hips leaves the improvement direction intact (median 33.4 \rightarrow 25.1 mm) but sharply inflates the spread ($\sigma_{\text{NIQR}} \approx 20$ mm), driven by a few recordings whose reference hips fail. Figure~\ref{fig-sameview-perinfant-slope} breaks these improvements out per infant: gains hold across almost the entire cohort.

On our prior study's benchmark, fine-tuning improves over the SAM-3D-Body base on reprojection error (39.8 \rightarrow 36.7 px) and geometric consistency (GC@10 0.31 \rightarrow 0.36). It remains well below off-the-shelf Sapiens on 2D metrics (reprojection 22.8 px, GC@10 0.82), as expected given that Sapiens is the distillation teacher. The Sapiens column carries our prior study's published values (\parencite{joshi_markerless_2026}); the SAM-3D-Body and Ours columns are computed through our pipeline over the same 11 infants.

\begin{table}
\centering
\caption[]{Accuracy on 11 held-out infants. \textit{Multi-view consistency} rows are means over infants; \textit{teacher agreement} rows are medians ($\sigma_{\text{NIQR}}$). The three PA-MPJPE rows are nested keypoint sets, each Procrustes-aligned per frame with scale on the keypoints it scores: \textbf{11 core joints} (shoulders, elbows, wrists, knees, ankles, neck), our primary 3D metric; \textbf{+ hips}, whose reference positions are less reliable in occluded infants; and \textbf{MHR-70}, the model's full output, of which 40 keypoints are fingers. MHR-70 is scored against a different triangulated reference from the other two and is not comparable with them. Sapiens produces no monocular 3D body (---). PCK and GC are higher-is-better, pixel error and PA-MPJPE (mm) lower-is-better; best value per row in \textbf{bold}.}

{\scriptsize
\setlength{\tabcolsep}{3pt}
\begin{tabular}{@{}lccc@{}}
\toprule
Metric & Sapiens & SAM-3D-Body & \textbf{Ours} \\
\midrule
\multicolumn{4}{@{}l}{\textit{Multi-view consistency}}\\
Reprojection (px) $\downarrow$   & \textbf{22.8} & 39.8 & 36.7 \\
GC@10 $\uparrow$                 & \textbf{0.82} & 0.31 & 0.36 \\
\addlinespace[2pt]
\multicolumn{4}{@{}l}{\textit{Teacher agreement}}\\
Body PCK@10 $\uparrow$            & --- & 0.216 (0.08) & \textbf{0.418 (0.09)} \\
Face PCK@10 $\uparrow$            & --- & 0.219 (0.03) & \textbf{0.422 (0.09)} \\
PA-MPJPE, 11 core joints (mm) $\downarrow$ & --- & 25.5 (5.4)  & \textbf{22.2 (3.6)} \\
PA-MPJPE, $+$ hips (13 kpt) (mm) $\downarrow$ & --- & 33.4 (20.8) & \textbf{25.1 (20.1)} \\
PA-MPJPE, MHR-70 (mm) $\downarrow$ & --- & 27.3 (10.1) & \textbf{23.7 (7.1)} \\
\bottomrule
\end{tabular}
}
\label{tbl-main}
\end{table}

\subsection{Shape--scale regularization}

The reported model masks the gradient on the shape and scale rows of the pose head's output layer. Training those rows instead, under an L2 prior $\lambda_{ss}$ swept from strong (0.05) through light (0.005) to none (0), does not improve accuracy: on the held-out reference the masked model is best or tied for best on every metric, and with no prior ($\lambda_{ss}=0$) body proportions inflate, most visibly the head, into shapes atypical of infants. The full sweep --- a combined accuracy table over all four settings at their final checkpoints together with the corresponding mesh overlays --- is reported in the accompanying Supplementary Material document (Table S1, Fig. S1), submitted alongside this paper.

\subsection{Training loss composition}

The objective is dominated by the 2D keypoint terms; the render-based dense terms
(surface normals, silhouette, depth) contribute little and do not converge.
Table~\ref{tbl-loss-composition} decomposes the converged in-loop objective into each
term's weighted contribution (weight $\times$ loss value) as a fraction of the total.

\begin{table}
\centering
\caption[]{Loss-term composition of the reported run (converged in-loop eval values; the weighted terms sum to the total loss). The keypoint terms dominate the objective; the dense render terms are a small, non-converging fraction.}
\label{tbl-loss-composition}
\begin{tabular}{p{\dimexpr 0.200\linewidth-2\tabcolsep}p{\dimexpr 0.200\linewidth-2\tabcolsep}p{\dimexpr 0.200\linewidth-2\tabcolsep}p{\dimexpr 0.200\linewidth-2\tabcolsep}p{\dimexpr 0.200\linewidth-2\tabcolsep}}
\toprule
Term & Value & Weight & Weighted & \% \\
\hline
$L_\text{kpt}$, major joints & 2.37 & 1.0 & 2.37 & 32.6 \\
$L_\text{kpt}$, head & 2.40 & 1.0 & 2.40 & 33.0 \\
$L_\text{kpt}$, body & 3.43 & 0.5 & 1.72 & 23.6 \\
$L_\text{kpt}$, face & 6.57 & 0.1 & 0.66 & 9.0 \\
$L_\text{normal}$ & 0.185 & 0.5 & 0.093 & 1.3 \\
$L_\text{silh}$ & 0.243 & 0.1 & 0.024 & 0.3 \\
$L_\text{depth}$ & 0.037 & 0.5 & 0.019 & 0.3 \\
\bottomrule
\end{tabular}
\end{table}

The keypoint terms together account for 98.1\% of the objective; the dense terms
(normals + silhouette + depth) sum to 1.86\%.

\section{Discussion}

Our prior study \parencite{joshi_markerless_2026} demonstrated that two recent foundation models for human pose estimation exhibit complementary strengths: SAM 3D Body produces joint angles for an MHR kinematic model with reasonable 3D accuracy, while Sapiens achieves more precise 2D joint position estimates but lacks 3D joint angle information. This work shows that these complementary strengths can be combined through label-free cross-model distillation. We distill the 2D fidelity of Sapiens 2 into SAM 3D Body through a render-in-the-loop objective, using only unannotated infant video. On eleven held-out infants with multi-view data to provide high quality reference measurements (173 recordings), the fine-tuned student shows statistically significant improvements in 2D agreement with the Sapiens reference along with scale-invariant 3D error (PA-MPJPE), narrowing the trade-off our prior study identified without requiring any manual annotation or marker-based ground truth.

Recovering whole-body 3D kinematics, rather than 2D keypoints alone, from ordinary video is a meaningful step toward scalable video-only assessment of early motor development, particularly in settings where marker-based capture is impractical and contraindicated \parencite{joshi_markerless_2026}. The quality of selective motor control --- the ability to move individual joints independently of one another --- is a clinically important indicator of motor impairment, and its emergence in infancy predicts later cerebral palsy \parencite{Cahill-Rowley2014EtiologyPalsy, Peyton2024BabyLongitudinal}. While the absolute improvement in 3D accuracy is a few millimetres, relative to the size of an infant this is about a 10\% reduction in Procrustes-aligned error (25.5 to 22.2 mm). Our goal is ultimately to classify movements, reproduce clinical scores, and find clinically meaningful patterns in movement; it remains to be seen what level of accuracy is required for accurate classification of pathological movements. Notably, kinematic features computed from 2D keypoints alone, using a pose estimator never tuned for infants, already predict clinician-assessed General Movements Assessment scores \parencite{segado_assessing_2024}. Recovering 3D biomechanical kinematics should nonetheless provide a viewpoint-invariant and anatomically interpretable representation on which such classifiers can be built.

A parallel line of our work concerns what the model outputs rather than how well it tracks. Mesh recovery returns vertices and keypoints posed on non-anatomical kinematic trees and not the biomechanical joint angles. In concurrent work we added a biomechanical head to the same SAM 3D Body backbone that regresses the joint angles and segment scales of a biomechanical model directly from one image \parencite{cotton_biomechanical_2026}. That head is supervised by in-loop optimized targets from an inverse-kinematics solver we developed for monocular finger tracking \parencite{cotton_monocular_fingers_2026} rather than by paired biomechanical labels, so like the distillation reported here it trains on unlabeled images. Combining the two is a natural next step: the distillation described in this paper adapts the mesh to infants, and biomechanical distillation on top of it would return infant movement as joint angles from which clinical measures can be computed directly.

This work has several limitations. The most pronounced is that while the keypoint accuracy improves, both in 2D and 3D, the shape of the mesh still differs from the infant's, particularly at the head. This may indicate the need to further tune the hyperparameters of our distillation. The non-keypoint-based losses contributed only 2\% to the final loss at convergence, despite Figure~\ref{fig-channels} showing substantial errors in the silhouette.  MHR's shape space was designed for adults and may not well represent the large head-to-body ratio, short limbs, and distinct limb-to-torso proportions of early development \parencite{Sciortino2017InfantProportions, huang2021invariant}, so even a well-distilled student is fitting baby movement onto an adult-shaped body. This adult-mean limitation motivates a different shape model. ANNY \parencite{bregier_anny_2025} is a natural candidate: a scan-free, fully differentiable parametric model whose interpretable age phenotype spans infants through elders. No foundation model yet predicts ANNY parameters with infant surface contours, so developing one is a natural future direction alongside further hyperparameter tuning.

While we test on a held-out dataset, our videos were largely obtained in a format designed for assessing motor performance, with infants supine and alone, and with unsettled movement minimized. Because of this, generalization to other contexts, such as crawling or being held, remains to be seen. A more fundamental limitation is the absence of ground-truth annotations. Manual keypoint labeling and physical marker-based motion capture, the gold standards for pose estimation evaluation, are both highly impractical and largely contraindicated in this population: affixing markers to preterm or clinically monitored infants is invasive, disruptive, and in many settings not permitted. We therefore follow our prior study \parencite{joshi_markerless_2026} in using multi-view triangulation as the next best alternative, which provides a high-quality reference for 3D accuracy \parencite{Roy2022OnEstimation}. However, triangulation is itself model-derived and carries its own error, particularly at frequently occluded joints such as the hips, so our accuracy estimates are relative to an imperfect reference rather than an absolute ground truth.

\subsection{Conclusion}

Taken together, these results show that adapting a 3D body model to a data-scarce infant population does not require infant pose labels or marker-based ground truth --- only a frozen teacher, unlabeled video, and a differentiable render-in-the-loop objective to connect the two. This work moves toward markerless tools that could one day support early motor assessment at scale from monocular video, in exactly the settings where such tools are needed most.

\section*{Acknowledgments}
\small
We would like to thank Imani Mann and Grace Hoo for their assistance with participant recruitment and scheduling, Shawana Anarwala and Kayan Abdou for their assistance with experimental setup and data collection, and Kunal Shah for his assistance with infrastructure.

Research reported in this publication was supported by the Eunice Kennedy Shriver National Institute Of Child Health \& Human Development of the National Institutes of Health under Award Number R21HD117074, and by the National Center for Advancing Translational Sciences of the National Institutes of Health under Award Number T32TR005124. The content is solely the responsibility of the authors and does not necessarily represent the official views of the National Institutes of Health.

\appendix

\section{Appendix}

\subsection{Shape/scale constraint: does training the readout help?}\label{sec-reg-ablation}

The reported model masks the gradient on the shape and scale rows of the pose head's output layer (Methods). Here we ask whether \textit{training} those rows --- under an L2 prior of weight $\lambda_{ss}$, swept from strong through none --- improves the fit.
This spans a spectrum from the masked model, through progressively weaker priors, to no prior at all.

\begin{figure}[!htbp]
\centering
\includegraphics[width=1\linewidth]{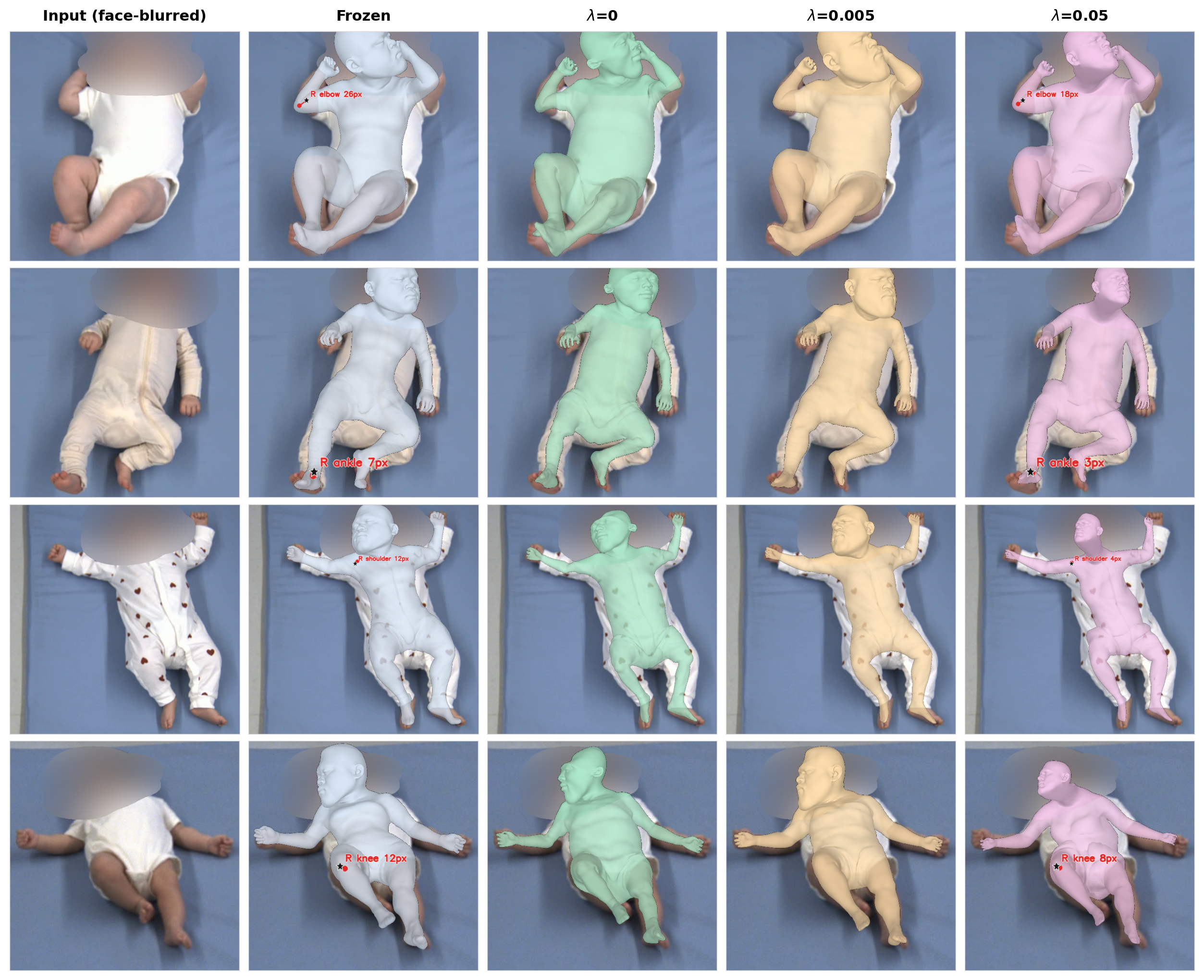}
\caption[]{Predicted MHR mesh overlaid on held-out infants across the shape/scale sweep: input, then the masked
model, $\lambda_{ss}=0$, $\lambda_{ss}=0.005$, and $\lambda_{ss}=0.05$. All meshes register to the same
person under the same camera. The unregularized run ($\lambda_{ss}=0$) produces a mixed set of shape changes: the torso becomes somewhat less narrow than in the reported model --- moving in the direction of infant proportions --- but the head simultaneously enlarges and rounds while the forehead narrows in an anatomically implausible way. The strongly regularized run and the reported model stay closer to the base model's proportions.}
\label{fig-reg-ablation-overlay}
\end{figure}

We compare four training runs --- the masked reported model and three trained-readout runs at
$\lambda_{ss}\in\{0.05,\,0.005,\,0\}$ --- each evaluated at its \textbf{final} checkpoint (no
loss-minimum selection), through the identical evaluation harness used in the main results, on the
same eleven held-out infants. Table~\ref{tbl-reg-ablation} reports the combined accuracy metrics and
Figure~\ref{fig-reg-ablation-overlay} shows the corresponding meshes on held-out infants. Both are
produced from the four final checkpoints by a single script that runs the evaluations, aggregates
the table, and renders the overlay figure end to end.

\begin{table}
\centering
\caption[]{Shape/scale-constraint ablation on the 11 held-out infants, each run evaluated at its FINAL checkpoint and ordered by decreasing shape/scale constraint (readout masked $\rightarrow$ readout trained with a vanishing shape regularizer $\lambda$). The three trained-readout runs also carry rebalanced dense-loss weights (normals, depth and silhouette raised by roughly 26$\times$, 128$\times$ and 100$\times$ relative to the masked run), so they vary the readout constraint and the dense weighting together rather than the constraint alone. Teacher-agreement (T) columns are the MEDIAN across infants of the per-camera 2D agreement with the Sapiens teacher (PCK@10px) and the hips-excluded core PA-MPJPE. Benchmark (B) columns are mean $\pm$ std over infants from the prior study's harness (non-facial reprojection error and GC@10; PA-MPJPE over the full MHR-70). PCK and GC are higher-is-better; pixel error and PA-MPJPE (mm) are lower-is-better; best value per column in bold.}

{\scriptsize
\setlength{\tabcolsep}{3pt}
\begin{tabular}{@{}llccccccc@{}}
\toprule
 & & & \multicolumn{3}{c}{Teacher agreement (T)} & \multicolumn{3}{c}{Benchmark (B)} \\
\cmidrule(lr){4-6} \cmidrule(lr){7-9}
Run & Constraint & $\lambda$ & Body & Face & PA-MPJPE & Reproj & GC@10 & MHR-70 \\
 & & & PCK@10 $\uparrow$ & PCK@10 $\uparrow$ & $-$hips (mm) $\downarrow$ & (px) $\downarrow$ & $\uparrow$ & (mm) $\downarrow$ \\
\midrule
SAM-3D-Body base & n/a & --- & 0.216 & 0.219 & 25.5 & 39.8 $\pm$ 5.7 & 0.31 $\pm$ 0.12 & 32.4 $\pm$ 13.7 \\
Masked (reported) & masked & n/a & \textbf{0.418} & \textbf{0.422} & \textbf{22.2} & \textbf{36.7 $\pm$ 3.8} & \textbf{0.36 $\pm$ 0.10} & \textbf{28.8 $\pm$ 11.4} \\
Trained, no reg & trained & 0 & 0.376 & 0.355 & 22.5 & 38.0 $\pm$ 4.4 & 0.33 $\pm$ 0.09 & 29.2 $\pm$ 9.8 \\
Trained, light reg & trained & 0.005 & 0.392 & 0.409 & 23.2 & 37.4 $\pm$ 4.7 & \textbf{0.36 $\pm$ 0.11} & 29.0 $\pm$ 10.8 \\
Trained, strong reg & trained & 0.05 & 0.342 & 0.227 & 29.4 & 40.7 $\pm$ 5.2 & 0.26 $\pm$ 0.08 & 43.2 $\pm$ 13.3 \\
\bottomrule
\end{tabular}
}
\label{tbl-reg-ablation}
\end{table}

This sweep is a negative result (Table~\ref{tbl-reg-ablation}): the masked model is best on every
held-out metric --- or tied for best, as it matches the light prior ($\lambda_{ss}=0.005$) on geometric consistency ---
and training those rows does not improve accuracy. The appeal of training those rows
is that the coefficients can deform the body toward whatever configuration best minimizes the pixel losses --- and with no
prior ($\lambda_{ss}=0$) this does reach the tightest in-loop (teacher-fidelity) fit --- but that
freedom is spent deforming \textit{shape} rather than improving 3D joint accuracy: the unregularized meshes
inflate the head and torso into  anatomically implausible shapes (Figure~\ref{fig-reg-ablation-overlay}) without
beating the masked model on the held-out reference. A strong prior ($\lambda_{ss}=0.05$, trained
twice as long) is worst of all; the light prior $\lambda_{ss}=0.005$ is the best of the trained-readout runs ---
matching the masked model on geometric consistency but trailing it on every other metric. We therefore report the masked model.
Its limitation is the counterpart to its strength: masking the readout keeps the meshes clean but still does not reach true
infant-specific body shape --- a limitation that joint-position metrics alone do not capture, and that motivates extending the method to
an infant-aware shape basis such as Anny \parencite{bregier_anny_2025}.

The in-loop loss measures fidelity to the teacher, not accuracy against the held-out reference. The
unregularized run ($\lambda_{ss}=0$) reaches the lowest in-loop loss of the three trained-readout runs and is
still not the best of them on the held-out metrics, let alone competitive with the masked model we
report. This mirrors the teacher-fidelity/accuracy gap discussed in the main text: a lower training
loss does not guarantee better held-out performance.

\end{document}